\documentclass{article}

     \PassOptionsToPackage{numbers, compress}{natbib}
\usepackage[final]{neurips_2019}

\usepackage[utf8]{inputenc} % allow utf-8 input
\usepackage[T1]{fontenc}    % use 8-bit T1 fonts
\usepackage{hyperref}       % hyperlinks
\usepackage{url}            % simple URL typesetting
\usepackage{booktabs}       % professional-quality tables
\usepackage{amsfonts}       % blackboard math symbols
\usepackage{nicefrac}       % compact symbols for 1/2, etc.
\usepackage{microtype}      % microtypography
\usepackage{graphicx}
\usepackage{times}
\usepackage{latexsym}
\usepackage{multirow}
\usepackage{amssymb, amsmath}
\usepackage{booktabs}
\usepackage{todonotes}

\newcommand{\fragment}{FLC }
\newcommand{\sentence}{SLC }

\newcommand{\fragmentend}{FLC}
\newcommand{\sentenceend}{SLC}

\newcommand{\del}[1]{}

\newcommand{\Ni}{({\em i})~}
\newcommand{\Nii}{({\em ii})~}

\title{Experiments in Detecting\\ Persuasion Techniques in the News}

\author{%
  Seunghak Yu\\
  MIT CSAIL\\
  Cambridge, MA, USA\\
  \texttt{seunghak@csail.mit.edu} \\
  \And
   Giovanni Da San Martino\\
    Qatar Computing Research Institute, HBKU \\
  Doha, Qatar \\
  \texttt{gmartino@hbku.edu.qa} \\
  \And
   Preslav Nakov \\
    Qatar Computing Research Institute, HBKU \\
  Doha, Qatar \\
  \texttt{pnakov@hbku.edu.qa} \\
}

\begin{document}

\maketitle

\begin{abstract}
    Many recent political events, like the 2016 US Presidential elections or the 2018 Brazilian elections have raised the attention of institutions and of the general public on the role of Internet and social media in influencing the outcome of these events. 
    We argue that a safe democracy is one in which citizens have tools to make them aware of propaganda campaigns. 
    We propose a novel task: performing fine-grained analysis of texts by detecting all fragments that contain propaganda techniques as well as their type. 
    We further design a novel multi-granularity neural network, and we show that it outperforms several strong BERT-based baselines. 
\end{abstract}

\section{Introduction} \label{sec:intro}

Journalistic organisations, such as \textit{Media Bias/Fact Check},\footnote{\url{http://mediabiasfactcheck.com/}} provide reports on news sources highlighting the ones that are propagandistic. 
Obviously, such analysis is time-consuming and possibly biased and it cannot be applied to the enormous amount of news that flood social media and the Internet. 
Research on detecting propaganda has focused primarily on classifying entire articles as propagandistic/non-propagandistic~\cite{AAAI2019:proppy,BARRONCEDENO20191849,rashkin-EtAl:2017:EMNLP2017}. 
Such learning systems are trained using gold labels obtained by transferring the label of the media source, as per \textit{Media Bias/Fact Check} judgment, to each of its articles. Such distant supervision setting inevitably introduces noise in the learning process \cite{Horne2018} and the resulting systems tend to lack explainability. 

We argue that in order to study propaganda in a sound and reliable way, we need to rely on high-quality trusted professional annotations and 
it is best to do so at the fragment level, targeting specific techniques rather than using a label for an entire document or an entire news outlet. 
Therefore, we propose a novel task: identifying specific instances of propaganda techniques used within an article. 
In particular, we design a novel multi-granularity neural network, and we show that it outperforms several strong BERT-based baselines.

\noindent Our corpus could enable research in propagandistic and non-objective news, including the development of explainable AI systems. A system that can detect instances of use of specific propagandistic techniques would be able to make it explicit to the users why a given article was predicted to be propagandistic. It could also help train the users to spot the use of such techniques in the news.

\section{Corpus Annotated with Propaganda Techniques} \label{sec:data}

We retrieved 451 news articles from 48 news outlets, both propagandistic and non-propagandistic according to \textit{Media Bias/Fact Check}, which professionals annotators\footnote{\url{http://www.aiidatapro.com}. The company performs professional annotations in the NLP domain, although they were not expert in propaganda techniques before this work.} annotated according to eighteen  persuasion techniques~\cite{Miller},  
ranging from leveraging on the emotions of the audience ---such as using \textit{loaded language} or \emph{appeal to authority}~\cite{Goodwin2011} and slogans~\cite{As2015}--- to using logical fallacies ---such as \textit{straw men}~\cite{Walton1996} (misrepresenting someone's opinion), hidden \textit{ad-hominem fallacies}, and \textit{red herring}~\cite[p.~78]{Weston2000} (presenting irrelevant data).\footnote{For a complete list see \url{http://propaganda.qcri.org/annotations/definitions.html}} 
Some of these techniques weren studied in tasks such as hate speech detection and computational argumentation \cite{habernal2018}. 

The total number of technique instances found in the articles, after the consolidation phase, is $7,485$, out of a total number of $21,230$ sentences (35.2\%). 
The distribution of the techniques in the corpus is also uneven: while there are $2,547$ occurrences of  \textit{loaded language}, there are only $15$ instances of \textit{straw man} (more statistics about the corpus can be found in~\cite{EMNLP19DaSanMartino}). 
We define two tasks based on the corpus described in Section~\ref{sec:data}: \Ni \textbf{\sentence (Sentence-level Classification)}, which asks to predict whether a sentence contains at least one propaganda technique, and \Nii \textbf{\fragment (Fragment-level classification)}, which asks to identify both the spans and the type of propaganda technique. 
Note that these two tasks are of different granularity, $g_1$ and $g_2$, namely tokens for \fragment and sentences for \sentenceend. 
We split the corpus into training, development and test, each containing 293, 57, 101 articles and 14,857, 2,108, 4,265 sentences, respectively. 

Our task requires specific evaluation measures that give credit for partial overlaps of fragments. 
Thus, in our precision and recall versions,
we give partial credit to imperfect matches at the character level, as in plagiarism detection~\citep{Potthast2010a}. 

Let $s$ and $t$ be two fragments, i.e., sequences of characters. We measure the overlap of two annotated fragments as  
$    C(s,t,h) = \frac{|(s\cap t)|}{h}\delta\left(l(s), l(t) \right)$, 
where $h$ is a normalizing factor, $l(a)$ is the labelling of fragment $a$, and $\delta(a,b)=1$ if $a=b$, and $0$ otherwise.

We now define variants of precision and recall able to account for the imbalance in the corpus:
\begin{equation}
P(S,T) = \frac{1}{|S|}\!\sum_{\begin{minipage}[t]{0.97cm}\footnotesize $s\in S,\\$ $t\in T$\end{minipage}}\!\!\! C(s,t,|s|), \hspace{1cm} 
R(S,T) = \frac{1}{|T|}\!\sum_{\begin{minipage}[t]{0.97cm}\footnotesize $s\in S,\\$ $t\in T$\end{minipage}}\!\!\! C(s,t,|t|), 
\label{eq:precisiontask3}
\end{equation}
In eq.~\eqref{eq:precisiontask3}, we define $P(S,T)$ to be zero if $|S|=0$ and $R(S,T)$ to be zero if $|T|=0$. 
Finally, we compute the harmonic mean of precision and recall in Eq.~\eqref{eq:precisiontask3} and we obtain an F$_1$-measure. 
Having a separate function $C$ for comparing two annotations gives us additional flexibility compared to standard NER measures that operate at the token/character level, e.g.,~we can change the factor that gives credit for partial overlaps and be more forgiving when only a few characters are wrong.

\section{Models} 
\label{sec:baselines}

We depart from BERT~\citep{devlin2018bert}, and we design three baselines.  

\begin{figure}[t]
\centering
\includegraphics[width=0.65\columnwidth]{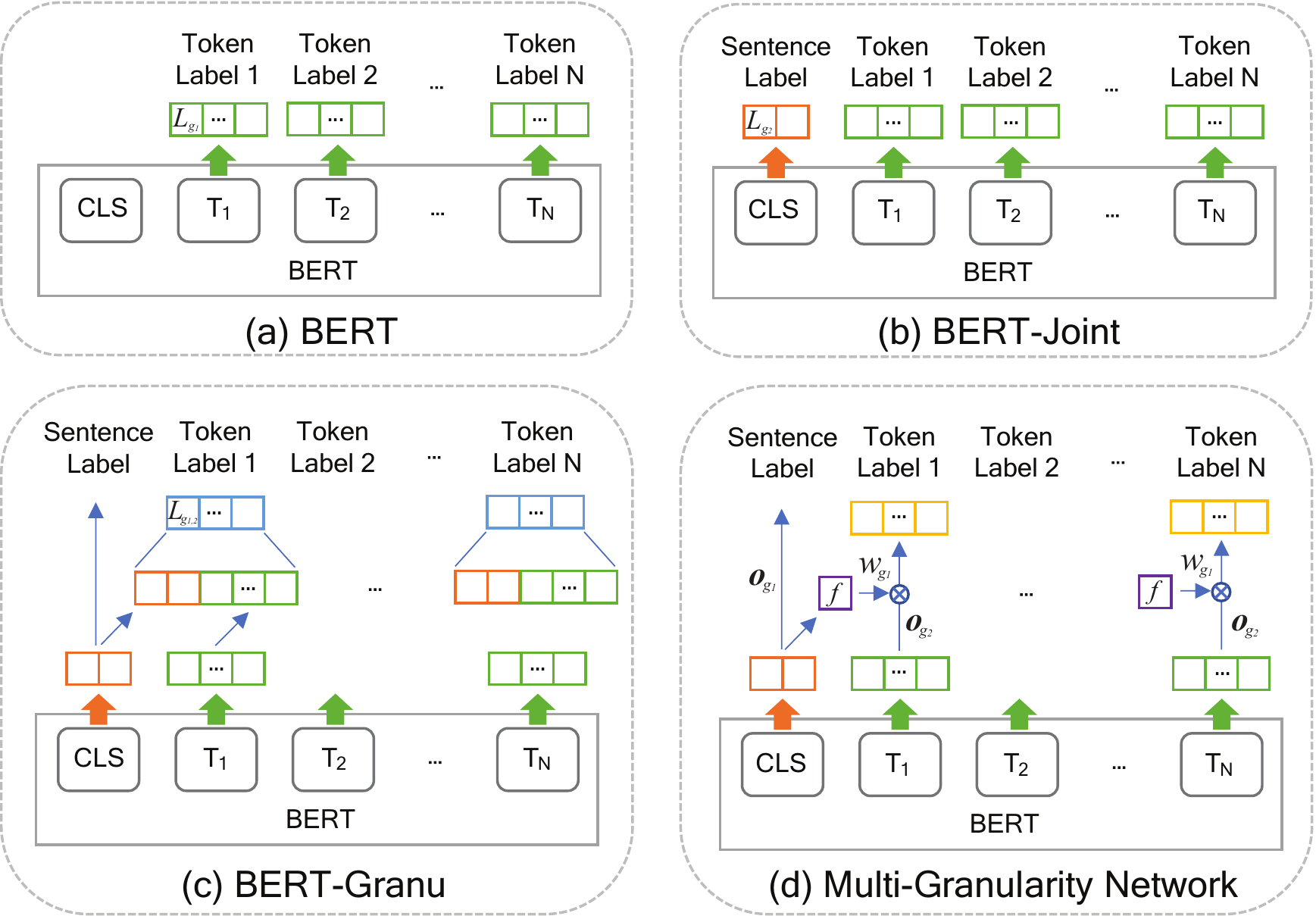}
\caption{The architecture of the baseline models (a-c), and of our multi-granularity network (d).}
\label{fig:arch}
\end{figure}

\textbf{BERT.} We add a linear layer on top of BERT and we fine-tune it, as suggested in~\cite{devlin2018bert}. For the \fragment task, we feed the final hidden representation for each token to a layer $L_{g_2}$ that makes a 19-way classification: does this token belong to one of the eighteen propaganda techniques or to none of them (cf.\ Figure~\ref{fig:arch}-a).
For the \sentence task, we feed the final hidden representation for the special \verb![CLS]! token, which BERT uses to represent the full sentence, to a two-dimensional layer $L_{g_1}$ to make a binary classification.

\textbf{BERT-Joint.} We use the layers for both tasks in the BERT baseline, $L_{g_1}$ and $L_{g_2}$, and we train for both \fragment and \sentence jointly (cf.\ Figure~\ref{fig:arch}-b).

\textbf{BERT-Granularity.} 
We modify BERT-Joint to transfer information from \sentence directly to \fragmentend. Instead of using only the $L_{g_2}$ layer for \fragmentend, we concatenate $L_{g_1}$ and $L_{g_2}$, and we add an extra 19-dimensional classification layer $L_{g_{1,2}}$ on top of that concatenation to perform the prediction for \fragment (cf.\ Figure~\ref{fig:arch}-c).

\textbf{Multi-Granularity Network.}
We propose a model that can drive the higher-granularity task (\fragmentend) on the basis of the lower-granularity information (\sentenceend), rather than simply using low-granularity information directly. Figure~\ref{fig:arch}-d shows the architecture of this model.

More generally, suppose there are $k$ tasks of increasing granularity, e.g., document-level, paragraph-level, sentence-level, word-level, subword-level, character-level.
Each task has a separate classification layer $L_{g_k}$ that receives the feature representation of the specific level of granularity $g_k$ and outputs $\boldsymbol{o}_{g_k}$.
The dimension of the representation depends on the embedding layer, while the dimension of the output depends on the number of classes in the task. The output $\boldsymbol{o}_{g_k}$ is used to generate a weight for the next granularity task $g_{k+1}$ through a trainable gate $f$:

\begin{equation}
w_{g_{k}} = f(\boldsymbol{o}_{g_k})
\end{equation}
The gate $f$ consists of a projection layer to one dimension and an activation function. The resulting weight is multiplied by each element of the output of layer $L_{g_{k+1}}$ to produce the output for task $g_{k+1}$:

\begin{equation}
\boldsymbol{o}_{g_{k+1}} = w_{g_{k}} * \boldsymbol{o}_{g_{k+1}}
\end{equation}
If $w_{g_{k}}=0$ for a given example, the output of the next granularity task $\boldsymbol{o}_{g_{k+1}}$ would be 0 as well. In our setting, this means that, if the sentence-level classifier is confident that the sentence does not contain propaganda, i.e.,~$w_{g_{k}}=0$, then $\boldsymbol{o}_{g_{k+1}}=0$ and there would be no propagandistic technique predicted for any span within that sentence. 
Similarly, when back-propagating the error, 
if $w_{g_{k}}=0$ for a given example, the final entropy loss would become zero, i.e.,~the model would not get any information from that example. As a result, only examples strongly classified as negative in a lower-granularity task would be ignored in the high-granularity task. 
Having the lower-granularity as the main task means that higher-granularity information can be selectively used as additional information to improve the performance, but only if the example is not considered as highly negative. 

For the loss function, we use a cross-entropy loss with sigmoid activation for every layer, except for the highest-granularity layer $L_{g_K}$, which uses a cross-entropy loss with softmax activation. Unlike softmax, which normalizes over all dimensions, the sigmoid allows each output component of layer $L_{g_k}$ to be independent from the rest. Thus, the output of the sigmoid for the positive class increases the degree of freedom by not affecting the negative class, and vice versa. As we have two tasks, we use sigmoid activation for $L_{g_1}$ and softmax activation for $L_{g_2}$. Moreover, we use a weighted sum of losses with a hyper-parameter $\alpha$:

\begin{equation}
\mathcal{L}_\mathcal{J} = \mathcal{L}_{g_1} * \alpha + \mathcal{L}_{g_2} * (1-\alpha) 
\end{equation}

Again, we use BERT~\cite{devlin2018bert} for the contextualized embedding layer and we place the multi-granularity network on top of it.

\section{Experiments and Evaluation}
\label{sec:experiments}

We used the PyTorch\footnote{\url{http://pytorch.org}} framework and the pretrained BERT model,\footnote{\url{http://github.com/huggingface/pytorch-pretrained-BERT}} which we fine-tuned for our tasks.\footnote{Our source code together with the dataset are available in GitHub: \url{http://anonymous.for.review}.}
To deal with class imbalance, we give weight to the binary cross-entropy according to the proportion of positive samples. For the $\alpha$ in the joint loss function, we use 0.9 for sentence classification, and 0.1 for word-level classification.
In order to reduce the effect of random fluctuations for BERT, all the reported numbers are the average of three experimental runs with different random seeds.
As it is standard, we tune our models on the dev partition and we report results on the test partition.

The left side of Table~\ref{Table:task3} shows the performance for the three baselines and for our multi-granularity network on the \fragment task. For the latter, we vary the degree to which the gate function is applied: using ReLU is more aggressive compared to using the Sigmoid, as the ReLU outputs zero for a negative input.
Table~\ref{Table:task3} (right) shows that using additional information from the sentence-level for the token-level classification (BERT-Granularity) yields small improvements. 
The multi-granularity models outperform all baselines thanks to their higher precision. 
This shows the effect of the model excluding sentences that it determined to be non-propagandistic from being considered for token-level classification. 

\begin{table}[t]
\centering
\begin{tabular}{@{}l@{\hspace{1mm}} c @{\hspace{1mm}}c@{\hspace{1mm}}c c@{\hspace{1mm}}c@{\hspace{1mm}}c@{}}
\toprule
 \multirow{2}{*}{Model} & \multicolumn{3}{c}{Task SLC} & \multicolumn{3}{c}{Task FLC} \\ 
 & P & R & F$_1$ & P & R & F$_1$ \\
 \midrule
 All-Propaganda  & 23.92 & 100.0 & 38.61 & - & -  & - \\
 BERT & \bf{63.20} & 53.16 & 57.74 & 21.48  & \bf 21.39 & 21.39\\
\,\,\, Joint      &  62.84 & 55.46 & 58.91 & 20.11 & 19.74 & 19.92\\
\,\,\, Granu      &  62.80 & 55.24 & 58.76   & 23.85 & 20.14 & 21.80\\
\multicolumn{2}{l}{\hspace{-3mm}Multi-Granularity} \\

\,\,\, ReLU    &  60.41 & \textbf{61.58} & \textbf{60.98} & 23.98 & 20.33 & 21.82\\
\,\,\, Sigmoid & 62.27 & 59.56 & 60.71 & \bf 24.42 & 21.05 & \bf 22.58\\
\bottomrule
\end{tabular}
\caption{Sentence-level (left) and fragment-level experiments (right).  \textit{All-propaganda} is a baseline that always output the propaganda class.\label{Table:task3}}
\end{table}

The right side of Table~\ref{Table:task3} shows the results for the \sentence task. We apply our multi-granularity network model to the sentence-level classification task to see its effect on low granularity when we train the model with a high granularity task. Interestingly, it yields huge performance improvements on the sentence-level classification result. Compared to the BERT baseline, it increases the recall by 8.42\%, resulting in a 3.24\% increase of the F$_1$ score. In this case, the result of token-level classification is used as additional information for the sentence-level task, and it helps to find more positive samples. This shows the opposite effect of our model compared to the \fragment task.

\section{Conclusions\label{sec:conclusions}}

We have argued for a new way to study propaganda in news media: by focusing on identifying the instances of use of specific propaganda techniques. Going at this fine-grained level can yield more reliable systems and it also makes it possible to explain to the user why an article was judged as propagandistic by an automatic system. 

We experimented with a number of BERT-based models and devised a novel architecture which outperforms standard BERT-based baselines. 
Our fine-grained task can complement document-level judgments, both to come out with an aggregated decision and to explain why a  document ---or an entire news outlet--- has been flagged as potentially propagandistic by an automatic system. 

In future work, we plan to include more media sources, especially from non-English-speaking media and regions. We further want to extend the tool to support other propaganda techniques.

\section{Acknowledgements}

This research is part of the Propaganda Analysis Project,\footnote{\url{http://propaganda.qcri.org}} which is framed within the Tanbih project.\footnote{\url{http://tanbih.qcri.org}} 
The Tanbih project aims to limit the effect of ``fake news'', propaganda, and media bias by making users aware of what they are reading, thus promoting media literacy and critical thinking. The project is developed in collaboration between the Qatar Computing Research Institute (QCRI), HBKU and the MIT Computer Science and Artificial Intelligence Laboratory (CSAIL).

\bibliography{emnlp-ijcnlp-2019,propaganda,other}
\bibliographystyle{abbrv}

\end{document}